\documentclass[10pt,twocolumn,letterpaper]{article}

\usepackage{cvpr}              % To produce the CAMERA-READY version
\usepackage[accsupp]{axessibility}  % Improves PDF readability for those with disabilities.
\usepackage{marvosym}

\definecolor{cvprblue}{rgb}{0.21,0.49,0.74}
\usepackage[pagebackref,breaklinks,colorlinks,allcolors=cvprblue]{hyperref}
\usepackage{amsthm,amsmath,amssymb}
\usepackage{tabularray}
\usepackage{subfiles}
\usepackage{mathrsfs}
\usepackage{float}
\def\paperID{30537} % *** Enter the Paper ID here
\def\confName{CVPR}
\def\confYear{2026}

\title{DINO-VO: Learning Where to Focus for Enhanced State Estimation}

\author{
Qi Chen$^{1,*}$ \quad
Guanghao Li$^{1,2,*}$ \quad
Sijia Hu$^{1}$ \quad
Xin Gao$^{1}$ \quad
Junpeng Ma$^{1}$ \quad \\
Xiangyang Xue$^{1}$ \quad 
Jian Pu$^{1, \text{\Letter}}$ \\
$^{1}$Fudan University \quad
$^{2}$Shanghai Innovation Institute \\
\tt\small{ \{qichen21, ghli22, sjhu23, gaoxin23, jpma24\}@m.fudan.edu.cn} \\ 
\tt\small{\{xyxue, jianpu\}@fudan.edu.cn} \\
\small{$^*$Equal contribution} 
}

\begin{document}
\maketitle
\begin{abstract}
We present DINO Patch Visual Odometry (DINO-VO), an end-to-end monocular visual odometry system with strong scene generalization. Current Visual Odometry (VO) systems often rely on heuristic feature extraction strategies, which can degrade accuracy and robustness, particularly in large-scale outdoor environments. DINO-VO addresses these limitations by incorporating a differentiable adaptive patch selector into the end-to-end pipeline, improving the quality of extracted patches and enhancing generalization across diverse datasets. Additionally, our system integrates a multi-task feature extraction module with a differentiable bundle adjustment (BA) module that leverages inverse depth priors, enabling the system to learn and utilize appearance and geometric information effectively. This integration bridges the gap between feature learning and state estimation. Extensive experiments on the TartanAir, KITTI, Euroc, and TUM datasets demonstrate that DINO-VO exhibits strong generalization across synthetic, indoor, and outdoor environments, achieving state-of-the-art tracking accuracy.
\end{abstract}
    
\section{Introduction}
\label{sec:intro}

Visual Odometry is a foundational technology in robotics and autonomous systems, closely intertwined with Simultaneous Localization and Mapping (SLAM). It enables agents to estimate their position while simultaneously understanding their environment. Over the decades, VO research has matured significantly, with monocular visual systems gaining attention due to their lightweight, low-cost nature. However, traditional monocular VO/SLAM systems~\cite{monoslamdavison2007, mur2015orb} often rely on manually designed modules, which suffer from poor cross-dataset generalization. 

Sparse feature-based monocular SLAM systems, such as ORB-SLAM3~\cite{orbslam3} and VPL-SLAM~\cite{vplslam}, offer robust mapping and localization performance in a variety of scenarios, thanks to their carefully crafted modules. However, these feature extraction methods often exhibit significant limitations, as they rely on heuristic techniques to assess feature importance during optimization. This reliance may lead to long-tail effects, making the system sensitive to hyperparameter changes and challenging to achieve consistent performance.

Other VO/SLAM systems that integrate learning-based feature extraction and matching modules have the same problem in sensitivity to hyperparameter tweaks. Since the learning-based feature extraction method is designed to enhance feature extraction matching ability, it is not directly responsible for downstream tasks such as state estimation. For instance, Light-SLAM\cite{lightslam} integrates SuperPoint\cite{superpoint} and LightGlue\cite{lightglue} for point feature extraction and matching and further enhances the accuracy of state estimation. Therefore, tuning the optimization module of the SLAM system, such as adjusting the edge weights during state estimation, remains necessary.

Recent advances in deep learning have led to the development of end-to-end learning-based VO/SLAM systems~\cite{droidslam, dpvo}, demonstrating improved cross-dataset generalization. End-to-end VO, such as DPVO\cite{dpvo}, achieves state-of-the-art performance in many complex scenarios. DPVO replaces all key components of the odometry system with learning-based methods and training learnable modules end-to-end and running with fewer hyperparameters. However, in DPVO\cite{dpvo}, plenty of patches extracted from the image frames were useless. DPVO's random patch selection strategy causes this problem since the random strategy might select patches in areas without effort-to-pose optimization, such as the sky or areas without context. To further improve the accuracy and efficiency of the odometry system, we build a novel odometry system that integrates learnable feature extraction, bridging the feature extraction and state estimation that can adaptively extract features that are more useful in bundle adjustment.

We introduce a novel end-to-end real-time deep-learning-based odometry called DINO Visual Odometry (DINO-VO). Our framework integrates a highly efficient and adaptive patch selection strategy and state estimation to enhance convergence even in complex environments. DINO-VO ensures improved state estimation accuracy and robustness by selecting only the most informative patches, particularly in feature-poor scenarios. Additionally, we incorporate the pre-trained monocular depth estimation model, Depth Anything v2\cite{depthanything}, which significantly boosts the quality of feature extraction by providing reliable depth priors. These depth priors help mitigate scale ambiguities and improve mapping accuracy and system stability.

Our contributions include:
\begin{itemize}
    \item A differential adaptive patch selector with a tailored training pipeline, improving state estimation accuracy
    \item A multi-task feature extractor based on a pre-trained vision model enhances cross-dataset generalization and extraction capabilities.
    \item Rigorous testing across indoor and outdoor datasets demonstrates that our SLAM system surpasses previous systems while retaining real-time efficiency.
\end{itemize}

\section{Related Works}

\subsection{Traditional VO/SLAM}
Traditional SLAM systems~\cite{mur2017orb, chen2024multi} have established a mature framework. As the first real-time monocular VSLAM algorithm, MonoSLAM~\cite{monoslamdavison2007} used Shi-Tomasi corners~\cite{shi1994good} for tracking in the frontend and employed an Extended Kalman Filter (EKF) for optimization in the backend. PTAM~\cite{PTAMKlein2007} divided VSLAM into two threads: mapping and tracking. The tracking thread used FAST corners~\cite{rosten2006machine} for pose estimation, while the mapping thread replaced the EKF with a nonlinear optimization algorithm. LSD-SLAM~\cite{lsdengel2014} used a direct method by optimizing pixel intensities and performed loop closure with feature points. The ORB-SLAM series~\cite{mur2015orb, mur2017orb, orbslam3} adopted PTAM’s dual-thread approach and introduced a loop closure detection thread. It used ORB features~\cite{orb} for tracking and DBoW~\cite{galvez2012bags} for loop closure detection, making it one of the most influential SLAM systems today. Besides, some systems~\cite{vplslam, xu2024airslam, li2025papl} employed line features to utilize the structural information in the environment. However, traditional SLAM relies on manually designed modules for state estimation, which imposes certain limitations on its generalizability across diverse scenarios.

\subsection{Learning-based VO/SLAM}
Learning-based SLAM combines the strong generalization capabilities of neural networks, introducing a new paradigm for SLAM-related research. Serval systems~\cite{tateno2017cnn, bloesch2018codeslam, li2020deepslam, czarnowski2020deepfactors, liftslam, lightslam} incorporated networks into the SLAM system to extract context feature of the image, predict depth, or optimize pose. While these systems produced remarkable results, the created modules were hand-crafted rather than end-to-end.

Towards end-to-end SLAM systems, DeepVO~\cite{wang2017deepvo} and UnDeepVO~\cite{li2018undeepvo} presented a novel end-to-end framework for monocular VO by using deep Neural Networks. DeepTAM~\cite{zhou2018deeptam} presented an entirely learned system for dense keyframe-based camera tracking and depth map estimation. Another end-to-end SLAM system utilized differential rendering techniques to produce a high-quality 3D map while inferring the pose via back-propropagation. NeRF~\cite{mildenhall2021nerf} based SLAM systems~\cite{li2026ec}, such as iMAP~\cite{sucar2021imap}, NICE-SLAM~\cite{zhu2022nice}, Co-SLAM~\cite{wang2023co}, ESLAM~\cite{johari2023eslam}, Go-SLAM~\cite{zhang2023go}, PLG-SLAM~\cite{plgSLAM}, and Loopy-SLAM~\cite{loopyliso}, used all kinds of map encodings to realize volumetric rendering. 3DGS~\cite{kerbl20233d} based SLAM systems~\cite{li2026artdeco, li2025constrained, deng2025best3dscenerepresentation, deng2025gaussiandwm3dgaussiandriving, deng2024compact}, such as SplaTAM~\cite{keetha2024splatam}, Gaussian Splatting SLAM~\cite{matsuki2024gaussian}, Photo-SLAM~\cite{huang2024photo}, and RTG-SLAM~\cite{peng2024rtg}, combined the fully differential feature of volumetric rendering and fast rasterization speed of 3DGS. Although these systems achieve end-to-end implementation, they overlook certain design aspects unique to SLAM in their end-to-end architecture.

End-to-end SLAM architecture with carefully designed modules~\cite{gaodeep, gaogood, Guo_2025_ICCV, hu2025mpcformerphysicsinformeddatadrivenapproach, lian2026finetuningenoughparallelframework, zhang2024sparsevlm, ma2025mmg, 11235592, he2026dynamicvggt, tang2026decoupling, tang2026causalvad, zhou2026spatialreward, ma2026gift} unique to SLAM has recently gained popularity. By using RAFT-like approaches~\cite{teed2020raft} to estimate the optical flow and a Dense Bundle Adjustment Layer to evaluate the pose difference, DROID-SLAM~\cite{droidslam} achieved high accuracy across multiple datasets, demonstrating a solid generalization. Furthermore, DPVO~\cite{dpvo} and DPV-SLAM~\cite{dpslam} improved upon DROID-SLAM~\cite{droidslam} by using sparse optical flow instead of dense optical flow, which conserves memory and increases processing speed while maintaining comparable accuracy. Nevertheless, the selected patches in the DPV~\cite{dpvo, dpslam} series do not contribute equally to bundle adjustment. Experiments reveal that among all patches specified in the DPV~\cite{dpvo, dpslam} series, only about 20\% make valuable contributions to pose estimation. Such low proportion is primarily due to the random patch selection strategy employed in the DPV~\cite{dpvo, dpslam} series. We implemented a self-attention selection mechanism using a multi-task feature extractor to identify and prioritize high-contribution patches.

\begin{figure*}[t]
    \centering
    \includegraphics[width=0.9\linewidth]{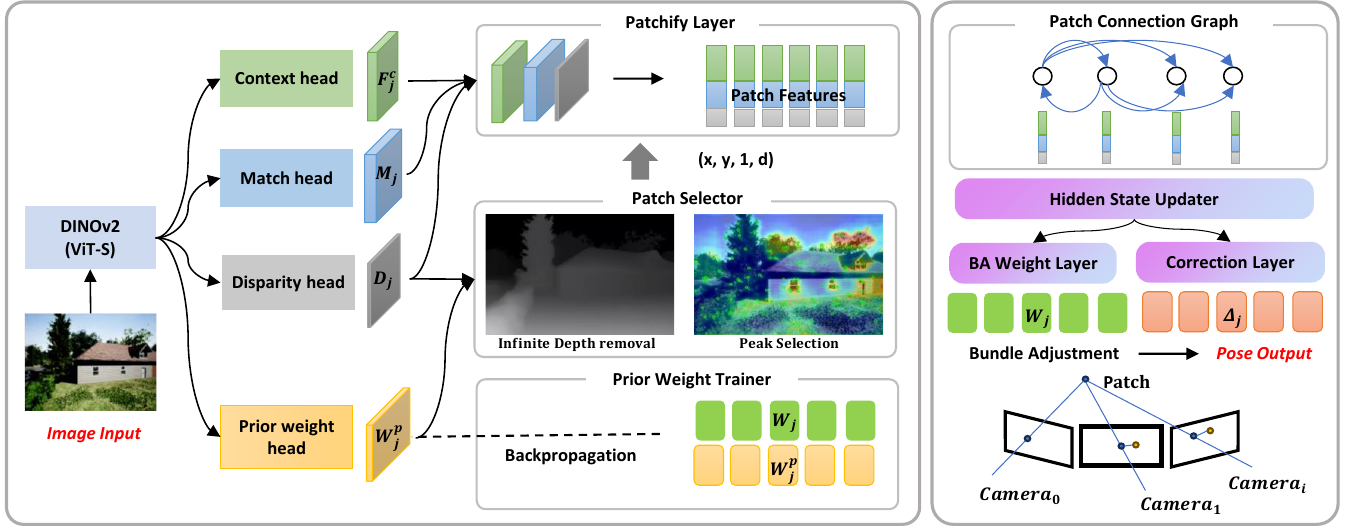}
    \caption{The overview of the system. Three modules establish our system from left to right: Multi-task Feature Extractor, Adaptive Patch Selector, and Sparse Bundle Adjustment Layer. The Multi-task Feature Extractor extracts the corresponding features for matching, selecting, and adjusting the bundle. The Adaptive Patch Selector selects high-weight patch features for bundle adjustment. The Sparse Bundle Adjustment Layer performs bundle adjustment to optimize the pose in the factor graph.}
    \label{fig:system}
\end{figure*}

\section{Method}
\label{sec:method}
  Given a set of sequential monocular images $ \{ I_i \}_{i=1}^{M} $ with known camera intrinsics $ K \in \mathbb{R}^{3 \times 3} $, we predicts camera poses $ \{ R_i|t_i \}_{i=1}^{M} $ and a sparse map. As illustrated in Fig.\ref{fig:system}, our system contains three main components: Multi-task Feature Extractor, Adaptive Patch Selector, and Sparse Bundle Adjustment Layer. 
   % similar to DPV-SLAM~\cite{dpslam}
   
\subsection{System Pipeline}
\label{sec:overview}
Initially, our multi-task feature extractor takes $j$-th image $\boldsymbol{I}_j$ with size $H * W$ as input and extracts the matching features $\boldsymbol{M}_j \in \mathbb{R}^{\frac{H}{4} \times \frac{W}{4}}$, context features $\boldsymbol{F}^c_j \in \mathbb{R}^{\frac{H}{4} \times \frac{W}{4}}$, inverse depth map $\boldsymbol{D}_j \in \mathbb{R}^{\frac{H}{4} \times \frac{W}{4}}$ and the prior weight map $\boldsymbol{W}^p_j \in \mathbb{R}^{\frac{H}{4} \times \frac{W}{4}}$. Subsequently, our adaptive patch selector extracts the candidate $N$ patches with size $p$ based on $\boldsymbol{W}^p_j$ and $\boldsymbol{D}_j$. The $l$-th patch among the total patches in $j$-th image, denoted as $\boldsymbol{P}_j^l = [\boldsymbol{x}, \boldsymbol{y}, \boldsymbol{1}, \boldsymbol{d}]^T$, contains pixel coordinates $\boldsymbol{x}, \boldsymbol{y},  \in \mathbb{R}^{1 \times p^2}$ in the features map and the inverse depth of patch $\boldsymbol{d} \in \mathbb{R}^{1 \times p^2}$  extracted from $\boldsymbol{D}_j$. 

We construct a bipartite patch graph $\mathscr{G}$ by connecting a patch $\boldsymbol{P}_j^l$ to every frame $i$ within a distance $r$ from $j$, each edge representing the projection $\boldsymbol{P}_{ji}^l$ of the original patch $\boldsymbol{P}_j^l$ onto the frame:
\begin{equation}
    \boldsymbol{P}^l_ {ji} \sim \boldsymbol{\bar{K}}\boldsymbol{T}_ {1} \boldsymbol{T}_ {j}^ {-1} \boldsymbol{\bar{K}}^ {-1} \boldsymbol{P}_ {j}^ {l} , \boldsymbol{\bar{K}}=
    \begin{pmatrix}
        \boldsymbol{K} & 0 \\
        0^T & 1
    \end{pmatrix},
\end{equation}
where $\boldsymbol{K}$ is the intrinsic matrix of the camera, $\boldsymbol{T}_i$ and $\boldsymbol{T}_j$ are the pose matrix of $i$-th and $j$-th image frame that transform from world coordinate to camera coordinate, we express the process as $\boldsymbol{P}^l_{ji} = \omega (\boldsymbol{T}_j, \boldsymbol{T}_i, \boldsymbol{P}_{j}^l)$.

Similar with  DPVO~\cite{dpvo}, we use matching features $\boldsymbol{M}_j$ to compute the correlation $\boldsymbol{C} \in \mathbb{R}^{p \times p \times 7 \times 7}$ between patch and its projection frame:

\begin{equation}
\mathbf{C}_{u v \alpha \beta}=\left\langle\mathbf{g}_{u v}, \mathbf{f}\left(\mathbf{P}_{k j}^{\prime}(u, v)+\Delta_{\alpha \beta}\right)\right\rangle,
\end{equation}
where $\mathbf{g}_{uv}$ is the matching feature of the patch extracted from $\boldsymbol{M}_j$ by patchify layer, $\mathbf{f}\left(\mathbf{P}_{ji}^{l}(u, v)+\Delta_{\alpha \beta}\right)$ is the matching feature of the patch in its projection frame. $\boldsymbol{\Delta}$ is to be a $7 \times 7$ integer grid centered at 0 indexed by $\alpha$ and $\beta$.

The extracted edges from the patches projections, patch context features $\boldsymbol{f}^c$ extracted from $\boldsymbol{F}^c$, and their corresponding correlation $\boldsymbol{C}$ are then fed into the hidden state updater to estimate a posterior weight $\boldsymbol{w}^{l}_{ji}$ for the sparse bundle adjustment and a 2D optical flow corrections $\boldsymbol{\Delta}^l_{ji}$ of $\boldsymbol{P}_{ji}^l$. Finally, the sparse bundle adjustment Layer optimizes the related poses.

\subsection{Multi-task Feature Extractor}
\label{sec:extractor}
\begin{figure}
    \centering
    \includegraphics[width=\linewidth]{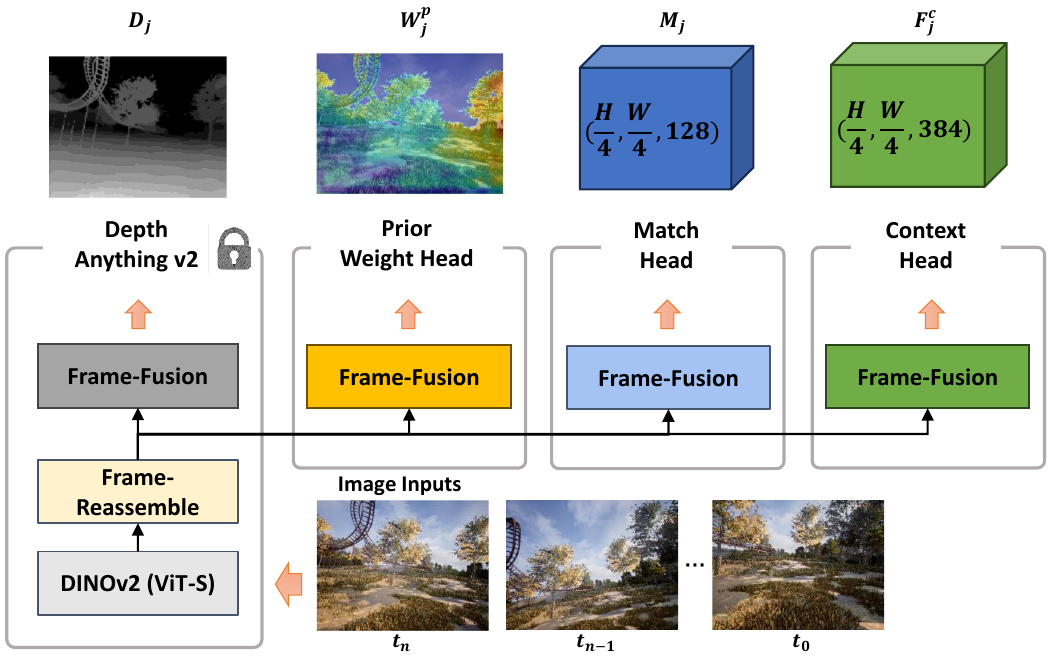}
    \caption{Architecture of the Multi-task Feature Extractor. The module predicts context features, matching features, an inverse depth map, and a prior weight map from a single RGB image.}
    \label{fig:mult-task}
\end{figure}

 As shown in Fig.\ref{fig:mult-task}, our multi-task feature extractor efficiently predicts context and matching features, inverse depth, and prior weight from a single RGB image using a single backbone, avoiding the overhead of using multiple backbones. In the SLAM system, an image's appearance and geometric features significantly impact subsequent tasks. We note that previous deep learning-based SLAM systems often lack the learning of scene geometry features (especially in monocular cases), relying solely on learned appearance features. However, simply integrating geometry prediction into our system is impractical because the whole model will take too much VRAM. Training four backbones to output the four features our system uses following the default train configuration in DPVO with sequences of length 15 needs almost 30 GB for batch size 1. Also, training geometry prediction models from the beginning in a single dataset has a domain shift problem. 
 
 Our feature extractor leverages a multi-task architecture based on Depth Anything v2~\cite{depthanything}. The ViT-S version of DINOv2\cite{oquab2024dinov2} serves as our feature extraction backbone, ensuring real-time performance in processing. This pre-trained model has been trained on millions of unlabeled images, yielding robust performance across diverse datasets. Notably, the ViT-S backbone in Depth Anything v2 can adequately extract information to predict all required features. Therefore, during training,  we freeze the Depth Anything v2~\cite{depthanything} model. Also, to minimize the training and inference time, all of our extractor heads in Fig.\ref{fig:system} are four Fusion layers and share the same Reassemble Layers of Depth Anything v2~\cite{depthanything}, rather than utilizing four full DPT heads~\cite{dpthead}.

\subsection{Adaptive Patch Selector}
\label{sec:selector}
% analysis 是不是应该是 improvement
In end-to-end SLAM systems, the importance of patches varies across different image regions for effective sparse bundle adjustment. Previous DPVO~\cite{dpvo} adopt a random patch selection strategy, resulting in over half of the patches' weights predicted by the hidden state updater being below 0.5 within $\mathscr{G}$. Although the random patch selection strategy can lead to good pose estimation results, it is ineffective enough for further improvement. To combine the end-to-end pipeline with the SLAM-specific technique, we propose a novel differentiable patch selection strategy to maximize the proportion of effective patches without relying on heuristic methods.

\subsubsection{Inference Phase}

\begin{figure}
    \centering
    \includegraphics[width=\linewidth]{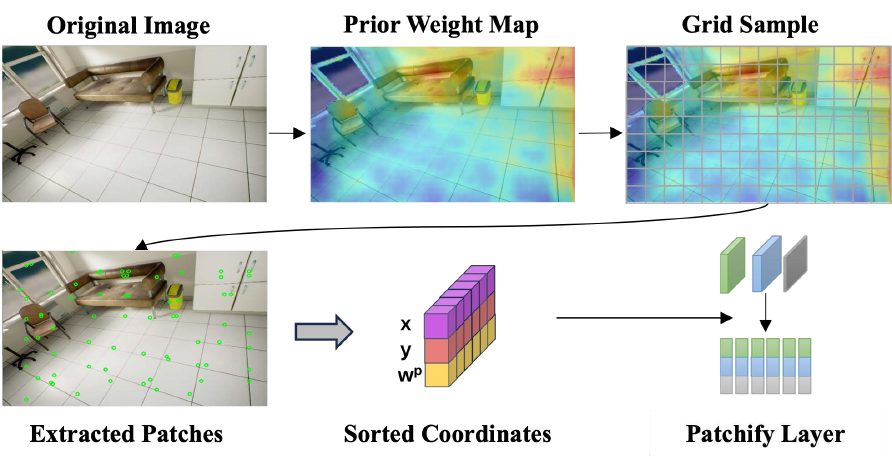}
    \caption{Pipeline for the adaptive patch selector. We utilize the prior weight and depth maps to uniformly select high-weight patches, which is more useful for the next sparse bundle adjustment.}
    \label{fig:patchSelector}
\end{figure}
During inference, the prior weight head predicts prior weight map $\boldsymbol{W}^p_j$ for frame $j$. It indicates the prior weight of all mapping patch coordinates extracted from the $j$-th image during bundle adjustment. We set the prior weight in $\boldsymbol{W}^p_j$ to $0$ if the corresponding pixel's depth is infinite:

\begin{equation}
    \boldsymbol{W}^{p'}_j = \boldsymbol{W}^p_j * \boldsymbol{mask},\ \boldsymbol{mask}_{uv} = \begin{cases} 
    1 & \text{if } \boldsymbol{D}_{uv} > 0 \\
    0 & \text{else }
    \end{cases},
\label{equ:combine}
\end{equation}
 $\boldsymbol{W}^{p'}_j$ is the post-processed prior map weight map, and $(u,v)$ is the pixel position of $\boldsymbol{D}$. Further, inspired by SuperPoint~\cite{superpoint}, we partition $\boldsymbol{W}^{p'}_j$ into n-by-n pixel regions,  extracting the position of the maximum prior weight in each region. These positions are sorted based on their weights, and the top $N$ positions are selected as our final patches.

\begin{figure*}
    \centering
    \includegraphics[width=0.9\linewidth]{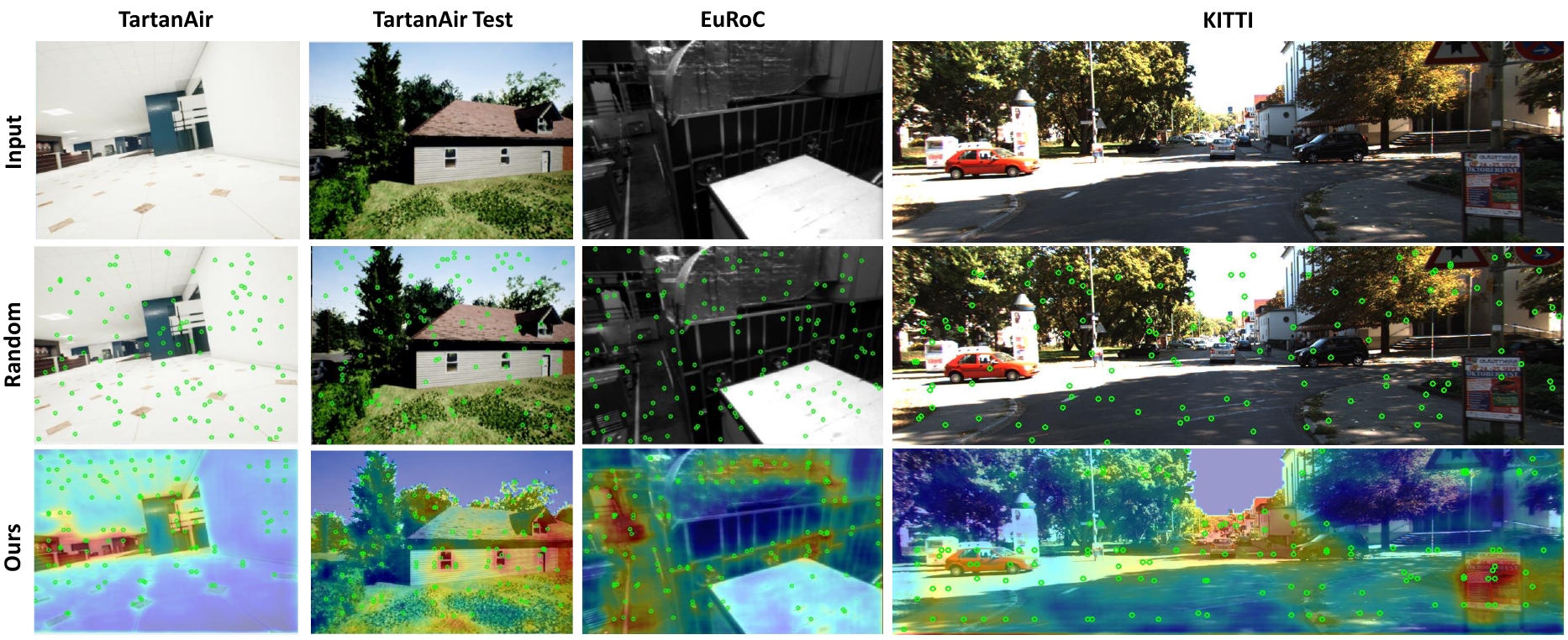}
    \caption{Comparison of our adaptive patch selector with existing systems. The first row is the input image from different datasets. The second row presents the random patch selection strategy proposed in \cite{dpvo} and \cite{dpslam}, while the third row illustrates our patch selection strategy, highlighting its focus on areas that contribute significantly to bundle adjustment.}
    \label{fig:patch_selection}
\end{figure*}

\begin{figure}
    \centering
    \includegraphics[width=0.90\linewidth]{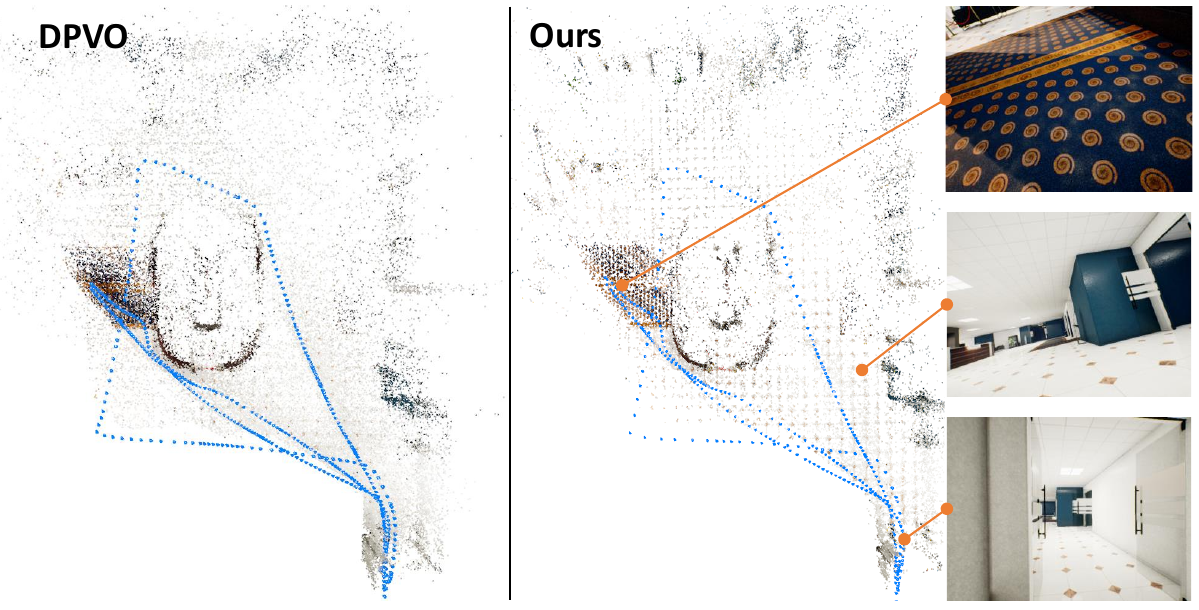}
    \caption{Comparison of reconstruction results on TartanAir~\cite{tartanair}. Our sparse map is more informative than DPVO's, particularly in the texture-rich area, as shown in the three small images on the right side.}
    \label{fig:mapping_results}
\end{figure}

 \subsubsection{Training by Distillation}
 \label{sec:distillation}
 Our prior weight head predicts every patch's prior weight in the subsequent sparse bundle adjustment for adaptively selecting high-weight patches. The hidden state updater also predicts a posterior weight for every patch. However, directly using the posterior weight to select high-weight patches is time-consuming cause every iteration using a hidden state updater needs to process a large factor graph. Reasonably, we perform distillation from the hidden state updater to our prior weight head while training the hidden state updater.
 
 Unlike the patch selection strategy in the inference phase,  we use a grid-random strategy to train our model. For $j$-th frame in the training phase, we divide all feature maps $\boldsymbol{F}^c_j$, $\boldsymbol{M}_j$, $\boldsymbol{W}_j$, $\boldsymbol{D}_j$ into m-by-m image regions and random extract the p-by-p patch $\boldsymbol{P}^l_j$ from these regions for mapping. Then we construct the bipartite patch graph $\mathscr{G}$. 
 
 For a patch $\boldsymbol{P}^l_{j}$, the initial ground truth for prior weight head, denoted as $\boldsymbol{w}^p_{l}$,  obtained from the max value in the posterior weight $\boldsymbol{w}_{l}$, which is the weight of $l$-th patch project to $j$-th frame (its original frame). To ensure the effectiveness and numbers of projections, we select $(j-1)$-th, $j$-th, and $(j+1)$-th frames' projections $\boldsymbol{P}^{(\cdot)}_{(j-1)j}$'s, $\boldsymbol{P}^{(\cdot)}_{jj}$'s and $\boldsymbol{P}^{(\cdot)}_{(j+1)j}$'s weights to form the $j$-th frame's ground truth $\boldsymbol{w}_j^{gt}$, where $(\cdot)$ represents all the projections in the original frame. Ground truth poses are used to compute the coordinates of the projections, allowing us to identify the corresponding pixel value in the predicted prior weight map, denoted as $\boldsymbol{w}_j^{pred}$. Due to the random selection strategy, plenty of weights in the initial ground truth tend to be near zero. Thus, we refine the ground truth by discarding zero weights and retaining one-tenth of the non-zero weights below 0.1 as the negative samples. The loss for training the prior weight head is then:
 
 \begin{equation}
     L_{pw} = \frac{1}{n}\sum_{i=1}^{n} |\boldsymbol{w}_i^{pred} - \boldsymbol{w}_i^{gt}| .
 \end{equation}

\subsection{Sparse Bundle Adjustment Layer}
\label{sec:Sparse bundle layer}
For every edge in the factor graph, we construct the following objective function:
\begin{equation}
    \sum_{(l, i) \in \mathscr{G}}\left\|\left[\hat{\mathbf{P}}_{j i}^{l}+\boldsymbol{\Delta}^l_{ji}\right]-\omega\left(\mathbf{T}_{i}, \mathbf{T}_{j}, \mathbf{P}_{j}^{l}\right)\right\|_{\boldsymbol{w}^l_{ji}}^{2},
\end{equation}
where $\boldsymbol{w}^l_{ji}$ represents the weights predicted by the hidden state updater. We optimize this function to find the most accurate pose for related frames.

The disparity head provides valuable prior knowledge of the inverse depth for all patches, enabling us to filter out patches with infinite depth before feeding them into bundle adjustment. This results in patches being initialized with a stable inverse depth, which aids bundle adjustment to converge to better results. Contrary to DPVO~\cite{dpvo}, which sets the inverse depth of the patch to the same constant value. We initialize the inverse depth of the patch according to Depth Anything v2's predicted values. However, initializing inverse depth directly by the value from Depth Anything v2 can lead to instability of odometry since the inverse depth predicted by the disparity head may not maintain a consistent scale. To address this issue, we set all inverse depths of the patches to 0.5 during the initialization. After the odometry finishes initialization, we rescale the initial inverse depth of the patch $d^l_{j}$ as follows:

\begin{equation}
    d_{j}^{l,rescale} = d^l_{j} * \frac{ \lvert \mathbb{D}_{j-1,j-2} \rvert * median(\mathbb{D}_j) }{ \sum_{d^l \in \mathbb{D}_{j-1,j-2}}{d^l} },
\end{equation}
where $\mathbb{D}_j$ is the depth of optimized patches observed from frame j and $\lvert \cdot \rvert$ represents the corresponding set's length.

This rescaling ensures that the inverse depths of patches converge towards a consistent and appropriate scale for further processing.

\begin{table*}
\small
\centering
\tabcolsep=0.09cm
\renewcommand\arraystretch{1.1}
\begin{tabular}{c|cccccccc|cccccccc|c} 
\hline
      Methods        & \begin{tabular}[c]{@{}c@{}}ME\\000\end{tabular} & \begin{tabular}[c]{@{}c@{}}ME\\001\end{tabular} & \begin{tabular}[c]{@{}c@{}}ME\\002\end{tabular} & \begin{tabular}[c]{@{}c@{}}ME\\003\end{tabular} & \begin{tabular}[c]{@{}c@{}}ME\\004\end{tabular} & \begin{tabular}[c]{@{}c@{}}ME\\005\end{tabular} & \begin{tabular}[c]{@{}c@{}}ME\\006\end{tabular} & \begin{tabular}[c]{@{}c@{}}ME\\007\end{tabular} & \begin{tabular}[c]{@{}c@{}}MH\\000\end{tabular} & \begin{tabular}[c]{@{}c@{}}MH\\001\end{tabular} & \begin{tabular}[c]{@{}c@{}}MH\\002\end{tabular} & \begin{tabular}[c]{@{}c@{}}MH\\003\end{tabular} & \begin{tabular}[c]{@{}c@{}}MH\\004\end{tabular} & \begin{tabular}[c]{@{}c@{}}MH\\005\end{tabular} & \begin{tabular}[c]{@{}c@{}}MH\\006\end{tabular} & \begin{tabular}[c]{@{}c@{}}MH\\007\end{tabular} & Avg    \\ 
\hline
ORB-SLAM3*~\cite{orbslam3}    & 13.61                                           & 16.86                                           & 20.57                                           & 16.00                                           & 22.27                                           & 9.82                                            & 21.61                                           & 7.74                                            & 15.44                                           & 2.92                                            & 13.51                                           & 8.18                                            & 2.59                                            & 21.91                                           & 11.70                                           & 25.88                                           & 14.38  \\
DROID-SLAM*~\cite{droidslam}   & \underline{0.17}                                            & \underline{0.06}                                            & 0.36                                            & 0.87                                            & 1.14                                            & 0.13                                            & 1.13                                            & \textbf{0.06}                                            & \textbf{0.08}                                            & \underline{0.05}                                            & \textbf{0.04}                                            & \textbf{0.02}                                            & \textbf{0.01}                                            & 0.68                                            & 0.30                                            & \textbf{0.07}                                            & 0.33   \\
DPVO~\cite{dpvo}          & \textbf{0.14}                                            & 0.11                                            & \underline{0.18}                                            & \underline{0.50}                                            & \textbf{0.39}                                            & \underline{0.12}                                            & \underline{0.34}                                            & 0.14                                            & 0.26                                            & \textbf{0.04}                                            & \textbf{0.04}                                            & \underline{0.06}                                            & 0.63                                            & \textbf{0.20}                                            & \textbf{0.10}                                            & \underline{0.09}                                            & \underline{0.21}   \\ 
\hline
Ours          & 0.22                                            & \textbf{0.05}                                            & \textbf{0.17}                                            & \textbf{0.17}                                            & \underline{0.51}                                            & \textbf{0.04}                                            & \textbf{0.33}                                            & \underline{0.08}                                            & \underline{0.24}                                            & \textbf{0.04}                                            & \underline{0.07}                                            & \underline{0.06}                                            & \underline{0.32}                                            & \underline{0.29}                                            & \underline{0.18}                                            & \underline{0.09}                                            & \textbf{0.18}       \\
\hline
\end{tabular}
\caption{Results on the TartanAir monocular test split. Results are reported as the ATE with scale alignment, in which \textbf{bold} and \underline{underline} represents the best and second, seperately. For our framework, we report the mean of 5 runs. Methods marked with (*) use global optimization/loop closure.}
\label{tab:tartan_test}
\end{table*}

\begin{table*}
\small
\centering
\tabcolsep=0.21cm
\renewcommand\arraystretch{1.1}
\begin{tabular}{c|ccccccccccc|c} 
\hline
Methods       & 00    & 01     & 02    & 03   & 04   & 05     & 06    & 07    & 08     & 09    & 10     & Avg    \\ 
\hline
ORB-SLAM3*~\cite{orbslam3}     & 8.27  & x      & \textbf{26.86} & \textbf{1.21} & \underline{0.77} & 7.91   & 12.54 & 3.44  & \textbf{46.81}  & 76.54 & \textbf{6.61}   & x      \\
LDSO*~\cite{gao2018ldso}          & 9.32  & 11.68  & \underline{31.98} & 2.85 & 1.22 & \textbf{5.10}   & 13.55 & 2.96  & 129.02 & \textbf{21.64} & 17.36  & \underline{22.42}  \\
DROID-SLAM*~\cite{droidslam}    & 92.10 & 344.60 & x     & 2.38 & 1.00 & 118.50 & 62.47 & 21.78 & 161.60 & x     & 118.70 & x      \\
DPV-SLAM++*~\cite{dpslam}    & \underline{7.79}  & 12.11  & 43.01 & 2.51 & 0.79 & \underline{5.41}   & \underline{11.07} & \underline{1.69}  & 109.97  & 76.64 & 13.33  & 25.85  \\ 
DPVO~\cite{dpvo}           & 113.21  & 12.69  & 123.4 & 2.09 & \textbf{0.68} & 58.96 & 54.78 & 19.26  & 115.90  & 75.10 & 13.63  & 53.61  \\ 
\hline
Ours & 100.39  & \textbf{7.50} & 111.48 & 2.07 & 1.41 & 48.30 & 50.18 & 17.80 & \underline{91.47} & \underline{68.62} & 10.53 &  46.32      \\
Ours* & \textbf{3.79}  & \underline{8.95} & 38.78 & \underline{2.00} & 1.44 & 9.14 & \textbf{10.85} & \textbf{1.28} & 92.24 & 68.83 & \underline{8.88}       &  \textbf{22.38}    \\
\hline
\end{tabular}
\caption{Results on the KITTI Odometry dataset. Results are reported as ATE with scale alignment. For our system, we report the mean of 5 runs. Methods marked with (*) use global optimization/loop closure. x indicates that the system does not converge on the specified scene.}
\label{tab:kitti}
\end{table*}

\subsection{Objective Function}
\label{sec:training}
In addition to the distillation training strategy in Sec.~\ref{sec:distillation}, our model also employs ground truth poses and optical flow to supervise the learning process. Specifically, we utilize pose supervision $L_{pose}$ and flow supervision $L_{flow}$ and an additional prior weight loss $L_{pw}$ for training model end-to-end. The final loss function is a weighted combination of these components:
\begin{equation}
    L = \lambda_{pose} L_{pose} + \lambda_{flow} L_{flow} + \lambda_{pw} L_{pw}.
    \label{equ:loss}
\end{equation}

% We introduce the overall system in Sec.\ref{sec:overview}. In Sec.~\ref{sec:extractor}, Sec.~\ref{sec:selector}, and Sec.~\ref{sec:Sparse bundle layer}, we introduce the three modules in Fig.~\ref{fig:system}. Furthermore, Sec.~\ref{sec:training} presents the details of training the end-to-end network.
 
%  The main contribution is the unified end-to-end framework with carefully designed SLAM modules: the multi-task Feature Extractor and the Adaptive Patch Optimizer. These modules are embedded to incorporate SLAM techniques into the end-to-end deep neural network. Challenges occur when complex modules are embedded into the end-to-end framework.
\begin{figure}
    \centering
    \includegraphics[width=0.99\linewidth]{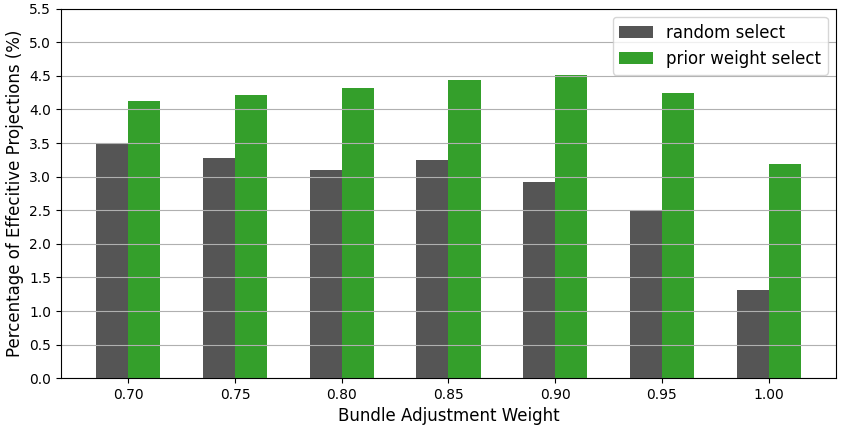}
    \caption{Percentage of effective patch projections in bundle adjustment. Our method selects more high-weight patches than the random selection method.}
    \label{fig:histogram}
\end{figure}

\section{Experiments}
\label{sec:experiments}
This section demonstrates that our system outperforms traditional and learning-based SLAM methods across four widely used datasets while maintaining a fast processing speed. Additional ablation studies further validate the efficacy of our approach.

\subsection{Experimental Setup}
\subsubsection{Baselines}
We benchmark our method against four state-of-the-art traditional SLAM methods: SVO~\cite{forster2014svo}, DSO~\cite{engel2017direct}, LDSO~\cite{gao2018ldso}, ORB-SLAM3~\cite{orbslam3}, and four state-of-the-art learning-based SLAM methods: TartanVO~\cite{wang2021tartanvo}, DROID-SLAM~\cite{droidslam}, DPVO~\cite{engel2017direct}, DPV-SLAM~\cite{dpslam}.

\subsubsection{Datasets and Metrics}
We evaluate our DINO-VO method on the TartanAir~\cite{tartanair}, TUM-RGBD~\cite{tartanair}, EuRoC~\cite{euroc}, and KITTI~\cite{kitti} benchmarks. We use the ATE RMSE (Absolute Trajectory Error Root Mean Squared Error) metric with scale alignment (in meters) for camera tracking,  evaluated via the EVO~\cite{grupp2017evo} tools. TartanAir, a synthetic dataset, provides a comprehensive training and testing set. The TUM and EuRoC datasets, consisting of indoor environments, allow us to assess our system’s performance in confined, complex settings. The KITTI dataset, featuring outdoor scenes, facilitates evaluation in large-scale environments. Despite being trained exclusively on synthetic data, our system achieves state-of-the-art performance on real-world datasets, demonstrating the cross-dataset generalizability of our approach.

\subsubsection{Implementation Details}
Our system runs on a desktop PC with 64G RAM, an Intel Core i7-12700KF CPU, and an NVIDIA RTX 3090 GPU. During training, the weights of $\lambda_{pose}, \lambda_{flow}, \lambda_{pw}$ for the objective function in Equ.\ref{equ:loss} is set to 10, 0.1, 10. The prior weight head, which learns from the hidden state updater's outputs during training iterations, starts its training process once the rest of the model converges. Specifically, we freeze the prior weight head for the first 5000 training iterations to ensure stability in training. Following the training pipeline outlined in \cite{dpvo}, we train the model for 240k iterations on a single GPU with a batch size of 1, using the AdamW optimizer with an initial learning rate of 8e-5, decaying linearly throughout the training process.

We run five trials during evaluation and report the mean results across each dataset. Our baselines include DPVO~\cite{dpvo} and its extended version DPV-SLAM~\cite{dpslam}. In the default configuration, our system extracts 100 patches per image and uses a 10-frame optimization window without employing proximity or the classical loop closure mechanism as proposed in \cite{dpslam}. The configuration incorporating proximity and classical loop closure is shown in the Ours* row of Tab.~\ref{tab:kitti} for large outdoor scenarios.

\begin{table*}
\centering
\small
\tabcolsep=0.36cm
\renewcommand\arraystretch{1.18}
\begin{tabular}{c|ccccccccc|c} 
\hline
Methods      & 360   & desk  & desk2 & floor & plant & room  & rpy   & teddy & xyz   & Avg    \\ 
\hline
ORB-SLAM3*~\cite{orbslam3}    & x     & \textbf{0.017} & 0.210 & x     & \underline{0.034} & x     & x     & x     & \textbf{0.009} & x      \\
DSO~\cite{engel2017direct}           & 0.173 & 0.567 & 0.916 & 0.080 & 0.121 & 0.379 & 0.058 & x     & 0.036 & x      \\
DROID-VO~\cite{droidslam}      & \underline{0.161} & 0.028 & \underline{0.099} & \textbf{0.033} & \textbf{0.028} & \underline{0.327} & \underline{0.028} & 0.169 & 0.013 & \underline{0.098}  \\
DPVO~\cite{dpvo}          & 0.179 & 0.026 & 0.113 & 0.061 & 0.040 & 0.469 & 0.040 & \textbf{0.083} & 0.011 & 0.114  \\ 
\hline
Ours          & \textbf{0.136} & \underline{0.021} & \textbf{0.052} & \underline{0.053} & 0.038 & \textbf{0.260} & \textbf{0.025} & \underline{0.131} & \underline{0.010} & \textbf{0.081}  \\
\hline
\end{tabular}
\caption{Results on the TUM-RGBD dataset. Results are reported as ATE with scale alignment. Results are reported as ATE with scale alignment. For our system, we report the mean of 5 runs. Methods marked with (*) use global optimization/loop closure. x indicates that the system does not converge on the specified scene.}
\label{tab:tum_rgbd}
\end{table*}

\begin{table*}
\centering
\small
\tabcolsep=0.23cm
\renewcommand\arraystretch{1.18}
\begin{tabular}{c|ccccccccccc|c} 
\hline
Methods      & MH01  & MH02  & MH03  & MH04  & MH05  & V101  & V102  & V103  & V201  & V202  & V203  & Avg    \\ 
\hline
Tartan VO~\cite{wang2021tartanvo}     & 0.639 & 0.325 & 0.550 & 1.153 & 1.021 & 0.447 & 0.389 & 0.622 & 0.433 & 0.749 & 1.152 & 0.680  \\
SVO~\cite{forster2014svo}           & 0.100 & 0.120 & 0.410 & 0.430 & 0.300 & 0.070 & 0.210 & x     & 0.110 & 0.110 & 1.080 & 0.294  \\
DSO~\cite{engel2017direct}           & \textbf{0.046} & \underline{0.046} & 0.172 & 3.810 & \underline{0.110} & 0.089 & \textbf{0.107} & 0.903 & \textbf{0.044} & 0.132 & 1.152 & 0.601  \\
DROID-VO~\cite{droidslam}      & 0.163 & 0.121 & 0.242 & 0.399 & 0.270 & 0.103 & 0.165 & 0.158 & 0.102 & 0.115 & \textbf{0.204} & 0.186  \\
DPVO~\cite{dpvo}          & 0.087 & 0.078 & 0.147 & 0.147 & 0.135 & \underline{0.048} & 0.137 & \underline{0.086} & 0.060 & \textbf{0.045} & 0.423 & \underline{0.127}  \\ 
\hline
Ours(random)  & 0.102 & \textbf{0.024} & \underline{0.116} & 0.143 & 0.166 & \textbf{0.047 }& \underline{0.116} &\textbf{ 0.036 }& 0.060 & \underline{0.059 }& 1.054 & 0.175 \\
Ours(w/o depth)  & 0.073 & 0.065 & 0.124 & \underline{0.140} & 0.160 & 0.060 & 0.142 & 0.368 & 0.058 & 0.090 & 0.743 & 0.184 \\
Ours          & \underline{0.056} & 0.053 & \textbf{0.094} &\textbf{ 0.124} & \textbf{0.109} & 0.068 & 0.190 & 0.101 & \underline{0.046} & 0.115 & \underline{0.290} & \textbf{0.113} \\
\hline
\end{tabular}
\caption{Results on the EuRoC dataset. Results are reported as ATE with scale alignment. For our system, we report the mean of 5 runs. x indicates that the system does not converge on the specified scene. random indicates using the random patch selection mecth as DPV-SLAM~\cite{dpslam} does, and w/o depth indocates we do not use the inverse depth map from our Multi-task Feature Extractor.}
\label{tab:euroc}
\end{table*}

\subsection{Results}
\subsubsection{Results on TartanAir Validation Split}
We use the 32-sequence validation split from DROID-SLAM~\cite{droidslam} and report the aggregated results. Our method achieves an AUC of 0.85, outperforming  DPVO~\cite{dpvo} with 0.8 and 0.71 DROID-SLAM in the [0, 1]m error window.

\subsubsection{Results on TartanAir Test Split}
Tab.\ref{tab:tartan_test} shows results on the TartanAir test set from the ECCV 2020 SLAM competition. Our system achieves the lowest average error across the previous method proposed in Tab.\ref{tab:tartan_test}. Compared with DPVO~\cite{dpvo}, our system improved accuracy relatively by 14\%. Notably, our system performs better in outdoor scenarios such as ME001, ME002, and ME003, which involve forest and town environments. Furthermore, Fig.\ref{fig:mapping_results} presents the 3D reconstruction results of DPVO and ours on TartanAir. Our adaptive patch selector enables our system to generate a sparse map that is more meaningful than the one produced by DPVO. For example, our system successfully reconstructs texture-rich floor patterns, whereas DPVO's sparse map, based on random patch selection, may lack the ability to capture such context patterns.

\subsubsection{Results on KITTI}

\begin{figure}
    \centering
    \includegraphics[width=0.8\linewidth]{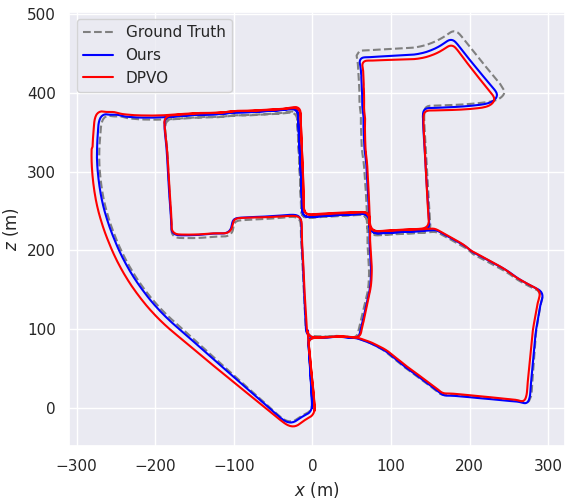}
    \caption{Trajector comparison between DPV-SLAM~\cite{dpslam} and our system with loop closure mechanism in KITTI Sequence 00}
    \label{fig:targetaryComaprsion}
\end{figure}

Tab.\ref{tab:kitti} presents results on the KITTI odometry dataset. Our system demonstrates state-of-the-art performance with an average ATE RMSE of 22.38m, particularly enhanced by incorporating loop closure (indicated by the asterisk $*$). With the help of our multi-task feature extractor and adaptive patch selector, our system performs better in large-scale outdoor areas (Sequence 00) and degenerated scenarios (Sequence 01). When comparing odometry results, our system outperforms DPVO~\cite{dpvo} by improving accuracy by 7.29m (DPVO's result is 53.61m).

Fig.\ref{fig:patch_selection} illustrates the qualitative results of our adaptive patch selector. Compared to DPVO~\cite{dpvo}, our system focuses more on meaningful objects, such as pipelines and buildings, than the sky. Besides, Fig.~\ref{fig:targetaryComaprsion} shows the trajectory comparison between our system and DPV-SLAM~\cite{dpslam} on sequence 00 of KITTI Odometry Dataset. Our system achieves a more accurate trajectory in specific segments, particularly in areas with high vehicle speeds and noisy, low-weight features (e.g., leaves). Our system tends to select higher-weight features, which are more effective for estimating optical flow changes.

\subsubsection{Results on TUM-RGBD}
Tab.\ref{tab:tum_rgbd} shows results on the TUM-RGBD dataset. Our method achieves the lowest average ATE RMSE of 0.081m, demonstrating consistently high accuracy across all nine tested sequences. Our system provides more reliable and accurate results in challenging conditions, such as blur and shake, compared to previous state-of-the-art SLAM methods. The consistently low ATE values highlight our method's robustness and adaptability to diverse environments.

\subsubsection{Results on EuRoC MAV}
Tab.\ref{tab:euroc} presents mapping results on the EuRoC dataset. Our method is accurate and consistent, achieving the lowest average error across all sequences. While DPVO~\cite{dpvo} and DROID-VO~\cite{droidslam} show reliable performance, they exhibit some variability across sequences. In contrast, systems like Tartan VO~\cite{wang2021tartanvo} and DSO~\cite{engel2017direct} demonstrate higher variability, affecting overall reliability. Our method's superior performance reflects its ability to handle various visual odometry challenges effectively.

\subsection{Ablations}
\subsubsection{Patch Selection}
To assess the effectiveness of our patch selection strategy, we statistically evaluate the weight of projections in bundle adjustment. In the EuRoC MH03 sequence, we collect all edge weights in the first one hundred frames. Subsequently, we showcase the percentage of high-weight projections in Fig.\ref{fig:histogram}. Our approach selects a higher percentage of high-weight patches than the random selection strategy. Additionally, as illustrated in Fig.\ref{fig:patch_selection}, our method focuses on areas with objects that are easier to track, such as pipelines, buildings, and other prominent structures.

To further evaluate the impact of patch selection on system accuracy, we conduct an ablation study on the EuRoC dataset. In this study, we replace the patch selection strategy described in Sec.\ref{sec:selector} with the random selection strategy in DPV-SLAM~\cite{dpslam}. The results of this modified system are shown in the Ours(Random) row in Tab.\ref{tab:euroc}. While the system performs well on most straightforward and moderate sequences, such as MH01, MH02, and MH03, it struggles with sequences involving rapid rotations, like V203. In these cases, the random patch selection strategy leads to system instability, causing a significant spike in ATE.

\subsubsection{Depth Prior}
To evaluate the influence of patches with inverse depth priors on system performance, we conduct an ablation in which we remove the fusion module that predicts inverse depth. Instead, we apply the prior weight map directly to the patch selector without post-processing.  For the bundle adjustment step, we follow DPVO~\cite{dpvo} by setting the initial inverse depth of patches to a constant value. We test this modified system on the EuRoC dataset, and the ATE is presented in the Ours(w/o depth) row in Tab.\ref{tab:euroc}. Although the modified system performs adequately on simpler sequences, its performance deteriorates on challenging sequences like V103 and V203, where the camera undergoes pure rotations without significant translation.

\section{Conclusion}
We introduce DINO-VO, an end-to-end visual odometry designed to overcome the limitations of previous systems. DINO-VO improves accuracy and efficiency by incorporating a learnable feature selection strategy directly tied to pose optimization, even in complex environments. Additionally, we integrate the pre-trained monocular depth estimation model Depth Anything v2, which enhances feature extraction and provides depth priors that further improve mapping accuracy and system stability. Our extensive experiments demonstrate that DINO-VO outperforms previous systems while maintaining real-time performance, highlighting its potential for robust and scalable SLAM in diverse real-world scenarios.

% In future work, we aim to incorporate a learnable loop closure module into our end-to-end SLAM system to improve robustness during loop closure further. Additionally, we plan to extend our monocular odometry system by fusing additional sensors, such as LiDAR or inertial measurement units (IMUs), to address the scale ambiguity inherent in monocular odometry.
% \input{sec/X_suppl}
{
    \small
    \bibliographystyle{ieeenat_fullname}
    \bibliography{main}
}

% WARNING: do not forget to delete the supplementary pages from your submission 

\end{document}